\documentclass[letterpaper, 10 pt, journal, twoside]{IEEEtran}
\IEEEoverridecommandlockouts

\DeclareUnicodeCharacter{0301}{\'{e}}

\usepackage[backend=biber,
            hyperref=true,
            url=false,
            isbn=false,
            doi=false,
            backref=false,
            style=ieee,
            citestyle=numeric-comp,
            sorting=nyt,%none
            block=none]{biblatex}

\usepackage[utf8]{inputenc}

\usepackage{float}
\usepackage{booktabs}
\usepackage{graphicx}
\usepackage{hyperref}
\usepackage{multirow}
\usepackage{balance}
\usepackage{placeins}
\usepackage{soul}
\usepackage{standalone}
\usepackage[font={footnotesize}]{caption}
\usepackage{subcaption}
\usepackage{tikz}
\usepackage{url}
\usepackage{amsmath, amssymb}
\usepackage{dsfont}
\usepackage[ruled]{algorithm2e}
\usepackage{algorithmic}
\usepackage{wrapfig}
\usepackage{soul}
\newcommand{\ourmethod}{\textsc{BUDS}}

\newcommand{\ourmethodimage}{\ourmethod-Image}
\newcommand{\ourmethodwsimage}{\ourmethod-WS-Image}
\newcommand{\ourmethodproprio}{\ourmethod-Proprio}

\newcommand{\tooluse}{\texttt{Tool-Use}}
\newcommand{\hammer}{\texttt{Hammer-Place}}
\newcommand{\kitchen}{\texttt{Kitchen}}
\newcommand{\multitask}{\texttt{Multitask-Kitchen}}
\newcommand{\realrobot}{\texttt{Real-Kitchen}}
\newcommand{\trainmulti}{\textbf{Train (Multi)}}
\newcommand{\trainsingle}{\textbf{Train (Single)}}
\newcommand{\test}{\textbf{Test}}

\newcommand{\task}[1]{\texttt{Task-{#1}}}
\newcommand{\variant}[1]{\texttt{Variant-{#1}}}
\newcommand{\revised}[1]{#1}% {\textcolor{red}{#1}}

\newcommand{\yifeng}[1]{{{#1}}}

\title{Bottom-Up Skill Discovery from Unsegmented Demonstrations for Long-Horizon Robot Manipulation}

\author{Yifeng Zhu$^{1}$, Peter Stone$^{1,2}$, Yuke Zhu$^{1}$
\thanks{Manuscript received: September 9, 2021; Revised December 6, 2021; Accepted Jan 10, 2022.}%Use only for final RAL version
\thanks{This paper was recommended for publication by Editor Markus Vincze upon evaluation of the Associate Editor and Reviewers' comments.
This work was partially supported by research grants from NSF, MLL Research Award, FLI, ARO, DARPA, Lockheed Martin, GM, and Bosch.} %Use only for final RAL version
\thanks{ $^{1}$Yifeng Zhu, Peter Stone, and Yuke Zhu are with the Department of Computer Science, the University of Texas at Austin. $^{2}$ Peter Stone is also with Sony AI.}
\thanks{Digital Object Identifier (DOI): see top of this page.}
}

\date{Sept 2021}

\begin{document}

\maketitle

\begin{abstract}
    We tackle real-world long-horizon robot manipulation tasks through skill discovery. We present a bottom-up approach to learning a library of reusable skills from unsegmented demonstrations and use these skills to synthesize prolonged robot behaviors. Our method starts with constructing a hierarchical task structure from each demonstration through agglomerative clustering. From the task structures of multi-task demonstrations, we identify skills based on the recurring patterns and train goal-conditioned sensorimotor policies with hierarchical imitation learning. Finally, we train a meta controller to compose these skills to solve long-horizon manipulation tasks. The entire model can be trained on a small set of human demonstrations collected within 30 minutes without further annotations, making it amendable to real-world deployment. We systematically evaluated our method in simulation environments and on a real robot. Our method has shown superior performance over state-of-the-art imitation learning methods in multi-stage manipulation tasks. Furthermore, skills discovered from multi-task demonstrations boost the average task success by $8\%$ compared to those discovered from individual tasks.\footnote{Project website with additional materials can be found at \url{https://ut-austin-rpl.github.io/rpl-BUDS}.}
\end{abstract}


% \keywords{ Skill Discovery, Robot Manipulation, Hierarchical Imitation Learning}
% \yifeng{Add Animesh's papers into citation}
\section{Introduction}
\label{sec:intro}
\IEEEPARstart{R}{eal}-world manipulation tasks challenge autonomous robots to reason about long-term interactions with the physical environment through the lens of raw perception. To tackle this challenge, temporal abstraction~\cite{sutton1999between,Precup00temporalabstraction} offers a powerful framework to model the compositional structures of manipulation tasks. The key idea is to use (sensorimotor) skills as the basic building blocks for synthesizing temporally extended robot behaviors. Recently, skill-based manipulation algorithms, including task and motion planning~\cite{kaelbling2011hierarchical,kaelbling2013integrated,garrett2021integrated} and hierarchical reinforcement learning~\cite{sutton1999between,dietterich1998maxq}, have demonstrated promising results in some restrictive settings. Nonetheless, it remains challenging to devise an effective yet automated way of building a rich repertoire of skills without costly manual engineering, especially when incorporating real-world sensory data.

\begin{figure}[t]
    \centering
    \includegraphics[width=.9\linewidth, trim=0cm 0cm 0cm 0cm,clip]{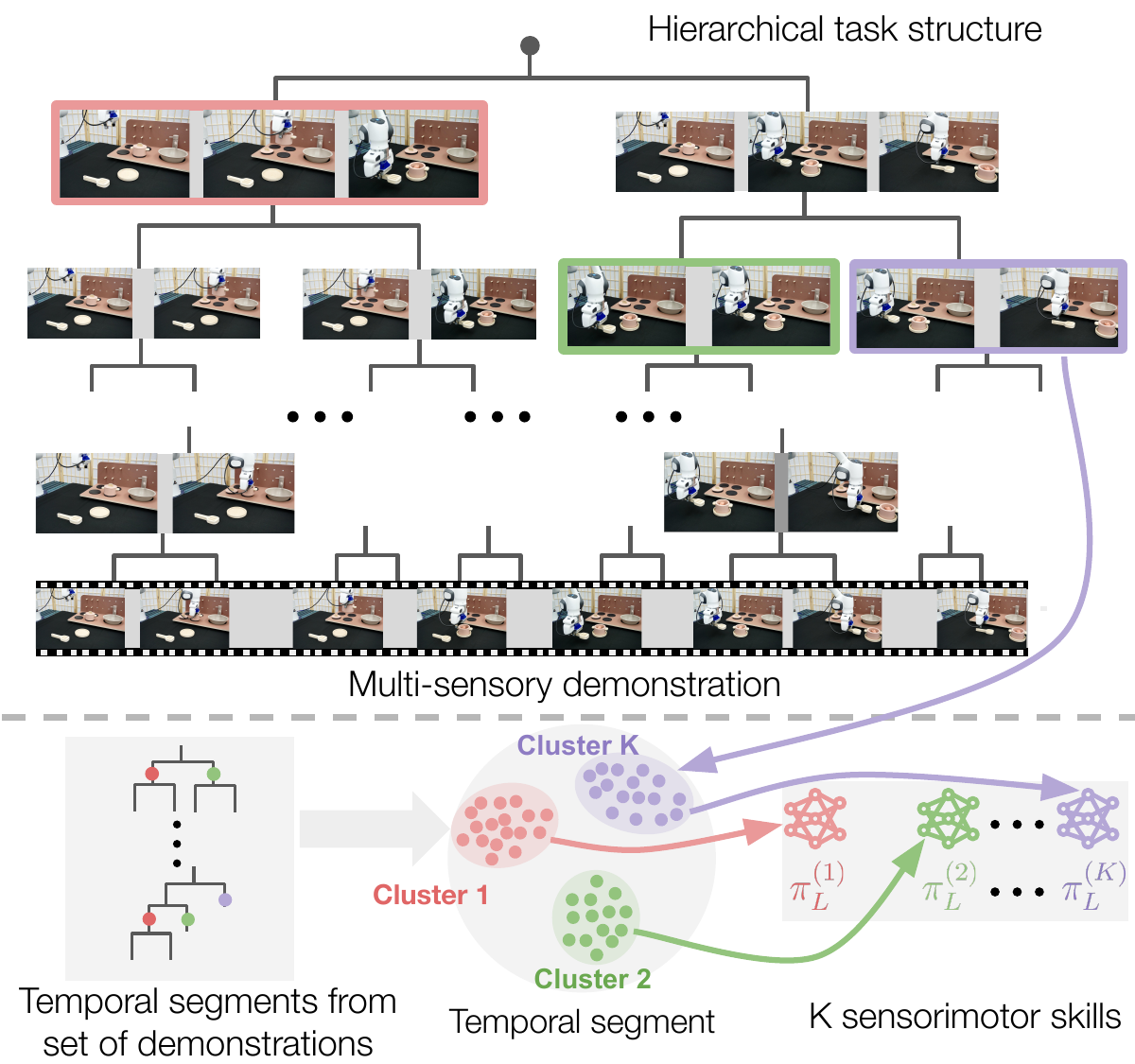}
    \caption{Overview of \ourmethod{}. We construct hierarchical task structures of demonstration sequences in a bottom-up manner, from which we obtain temporal segments for discovering and learning sensorimotor skills.}
    \label{fig:approach}
    \vspace{-7mm}
\end{figure}

Several paradigms have been investigated for automating skill acquisition. Notable ones include the \textit{options} framework~\cite{vezhnevets2017feudal,fox2017multi,gregor2016variational,konidaris2009skill,bagaria2019option,kumar2018expanding} and unsupervised skill discovery based on information-theoretic metrics~\cite{eysenbach2018diversity,hausman2018learning,sharma2019dynamics}. 
While initial successes have been achieved for discovering skills from a robot's self exploration, primarily in simulation, 
these methods exhibit a prohibitively high sample complexity or require access to ground-truth physical states, hindering their applicability on real robots.
An alternative strategy is to extract skills from human demonstrations, easing the exploration burdens. A direct approach is to manually annotate the demonstrations into coherent segments as skills~\cite{hovland1996skill}, but manual annotations of temporal segments can be ambiguous and costly. Instead, our approach discovers skills from a small set of \textit{unsegmented} human demonstrations with no temporal labels. It reduces the human efforts and improves the scalability.
Prior work on learning from unsegmented demonstrations used Bayesian inference~\cite{konidaris2009skill,niekum2012learning}, generative modeling~\cite{shankar2020learning,tanneberg2021skid,kipf2018compositional}, and dynamic programming~\cite{shiarlis2018taco} to temporally decompose the demonstrations into a sequence of shorter skill segments. However, these methods usually fall short of processing high-dimensional sensor data. Recent advances in hierarchical imitation learning~\cite{mandlekar2020iris,gupta2020relay,mandlekar2020learning}, modeled with deep neural networks, have shown promise in tackling long-horizon manipulation tasks in a hierarchical manner. They have focused on solving individual tasks rather than discovering a library of reusable skills across tasks.

\yifeng{We} develop a hierarchical approach to tackling real-world robot manipulation through skill discovery from unsegmented demonstrations. Our method \ourmethod{} (\textbf{B}uttom-\textbf{U}p \textbf{D}iscovery of sensorimotor \textbf{S}kills), presented in Figure~\ref{fig:approach}, starts with an unsupervised clustering-based segmentation model that extracts a library of sensorimotor skills from human demonstrations collected through teleoperation. Each skill is modeled as a goal-conditioned sensorimotor policy that operates on raw images and robot proprioceptiion. \ourmethod{} further learns a high-level meta controller that selects a skill and predicts the subgoal for the skill to achieve at any given state. Both the skills and the meta-controllers are trained with imitation learning on human demonstrations. Together it presents a scalable framework for solving complex manipulation tasks from raw sensory inputs, amendable to real-robot deployment.

Four key properties of \ourmethod{} are crucial for its effectiveness: 1) it uses bottom-up agglomerative clustering to build hierarchical task structures of demonstrations. These hierarchical representations offer flexibility for the imitation learner to determine the proper granularity of the temporal segments. 2) \ourmethod{} segments the demonstrations based on multi-sensory cues, including multi-view images and proprioceptive features. It takes advantage of the statistical patterns across multiple sensor modalities to produce more coherent and compositional task structures than using a single modality. 3) \ourmethod{} extracts skills from human demonstrations on multiple tasks that achieve different manipulation goals, facilitating knowledge sharing across tasks and improve the reusability of the discovered skills. 4) we train our goal-conditioned skills with the recently developed Hierarchical Behavior Cloning algorithm~\cite{mandlekar2020iris,gupta2020relay,mandlekar2020learning,tung2020learning}, producing perceptually grounded yet versatile skills for composition.

 We systematically evaluate our method in simulation and on a real robot and perform ablation studies to validate our design choices. \ourmethod{} achieves an average $66$\% success rate on three challenging vision-based manipulation tasks, \yifeng{which} the recent hierarchical imitation learning algorithms~\cite{mandlekar2020learning, gupta2020relay} struggled to solve. It also outperformed the most competitive baselines over $20\%$. We further show that skills learned from multi-task demonstrations boost the success rate by $8$\% compared to those learned for each task separately. Moreover, we show that the skills can be reused for solving new task variants that require different subtask combinations. Finally, we deploy \ourmethod{} on a real robot for a complex kitchen task, achieving a $56$\% success rate on par with our simulation results. For all experiments, \ourmethod{} is trained on 50-120 demonstrations for each task collected within 30min.

We summarize the three key contributions of this work: 1) We present a bottom-up clustering algorithm to discover sensorimotor skills from unsegmented demonstrations; 2) We introduce a hierarchical policy learning method that composes the skills for long-horizon, vision-based robot manipulation tasks; 3) We show the practical advantages of ~\ourmethod{} both in simulation and on real hardware.
\vspace{-0.2cm}

\section{Related Work}
\label{sec:related_works}

\noindent
\textbf{Robot Skill Discovery.} Skill discovery has been studied in a large body of existing works. A major line of works focuses on acquiring skills from self-exploration in environments. Many works fall into the options framework~\cite{sutton1999between}, discovering skills through hierarchical reinforcement learning~\cite{vezhnevets2017feudal,fox2017multi, gregor2016variational,konidaris2009skill,bagaria2019option,kumar2018expanding}. Other works use information-theoretic  metrics to discover skills from unsupervised interaction~\cite{eysenbach2018diversity,hausman2018learning,sharma2019dynamics}. These works typically require high sample complexity and operate on ground-truth physical states, hindering their applicability to real robot hardware. An alternative to self-exploration is to segment skills from human demonstrations, such as Bayesian inference~\cite{niekum2012learning, konidaris2010constructing, niekum2015online} and trajectory reconstruction~\cite{shankar2020learning, tanneberg2021skid}. These approaches produce temporal segmentation on low-dimensional physical states, difficult to scale to raw sensor data. 
Weakly supervised learning methods discover skill segments through temporal alignment on demonstrations~\cite{shiarlis2018taco, pirk2020modeling}, but require manual human annotations of task sketch.
Our work resonates with these works on skill discovery from human demonstrations; however, it directly operates on raw sensor data and \yifeng{requires} no manual labeling on execution stages in demonstrations. Similar to prior works~\cite{su2018learning, chu2019real}, we take advantage of multi-sensory cues in demonstrations. An important difference is that our method produces closed-loop sensorimotor policies, while the others focus primarily on learning task structures.

\noindent
\textbf{Bottom-up Methods in Perception and Control.} Bottom-up processing of sensory information traces back to Gibson's theory of direct perception~\cite{gibson1966senses}, of which the basic idea is that the higher level of information is built up on the retrieval of direct sensory information. Bottom-up methods have been successfully employed in various perception tasks. These methods construct hierarchical representations by grouping more fine-grained visual elements, such as pixels/superpixels for image segmentation~\cite{farag2016bottom} and spatio-temporal volumes for activity understanding~\cite{lan2015action,sarfraz2019efficient,sarfraz2021temporally}. Recently, bottom-up deep visual recognition models~\cite{newell2016stacked, law2018cornernet, zhou2019bottom} achieved competitive performances compared to the mainstream top-down methods.
The bottom-up design principles have also been studied for robot control. A notable example is the subsumption architecture developed in behavior-based robotics, which decomposes a complex robot behavior into hierarchical layers of sub-behaviors~\cite{brooks1986robust,brooks1991intelligence, nicolescu2002hierarchical, krishnan2017transition}. Our work leverages a similar bottom-up principle to discover hierarchical representations of human demonstrations. Furthermore, we demonstrate how imitation learners can exploit such hierarchies to scale to long-term manipulation behaviors.

\noindent
\textbf{Hierarchical Imitation in Robot Manipulation.} We leverage hierarchical imitation learning~\cite{le2018hierarchical} for learning policies of sensorimotor skills. Hierarchical imitation learning is a class of approaches which uses temporal abstractions to tackle longer-horizon tasks than vanilla imitation models. In particular, we use hierarchical behavior cloning~\cite{mandlekar2020iris,gupta2020relay,mandlekar2020learning,tung2020learning}, which recently shows great promises in robot manipulation. These methods learn a hierarchical policy where a high-level policy predicts subgoals and a low-level policy computes actions to achieve the subgoals, where the subgoals can be obtained through goal relabelling~\cite{andrychowicz2017hindsight}. One-shot imitation learning is a meta-learning framework that aims to learn from a single demonstration of test tasks~\cite{duan2017one, smith2019avid,yu2018one,xu2018neural}. Our work differs from prior work in that we extract a set of skills from multi-task demonstrations for task composition and directly handle raw sensory data.

\section{Approach}
\label{sec:approach}
We introduce \ourmethod{}, an approach for sensorimotor skill discovery and hierarchical imitation learning of closed-loop sensorimotor policies in robot manipulation. The core idea is to discover a set of reusable sensorimotor skills from multi-task, multi-sensory demonstrations, which can be composed to solve long-horizon tasks. An overview of \ourmethod{} is illustrated in Figure~\ref{fig:approach}. In the following, we first formalize our problem, and then present the two key steps of our approach: 1) skill segmentation with hierarchical agglomerative clustering on unsegmented demonstrations, and 2) learning skills and meta-controllers with hierarchical behavioral cloning.

\subsection{Problem Formulation}
\vspace{-1.5mm}
We formalize the problem of solving a robot manipulation task as a discrete-time Markov Decision Process $\mathcal{M}=(\mathcal{S}, \mathcal{A}, \mathcal{P}, R, \gamma, \rho_{0})$ where $\mathcal{S}$ is the state space, $\mathcal{A}$ is the action space, $\mathcal{P}(\cdot|s, a)$ is the stochastic transition probability, $R(s, a, s')$ is the reward function, $\gamma \in [0, 1)$ is the discount factor, and $\rho_{0}(\cdot)$ is the initial state distribution. Our goal is to learn a sensorimotor policy $\pi: \mathcal{S}\rightarrow\mathcal{A}$ that maximizes the expected return $\mathbb{E}[\sum^\infty_{t=0}\gamma^t R(s_t, a_t, s_{t+1})]$. In our context, $\mathcal{S}$ is the space of the robot's sensor data including raw images and proprioceptions, $\mathcal{A}$ is the space of the robot's motor actions, and $\pi$ is a closed-loop sensorimotor policy that we deploy on the robot to perform the task.

To tackle long-horizon tasks, we factorize the policy $\pi$ with a two-level temporal hierarchy. The low level consists of a set of $K$ skills, $\{\pi^{(1)}_L, \pi^{(2)}_L, \ldots, \pi^{(K)}_L\}$, each of which corresponds to a goal-conditioned policy $\pi^{(k)}_L: \mathcal{S}\times \Omega\rightarrow \mathcal{A}$, where $\omega\in\Omega$ is a vector that represents a goal, and $k$ is the skill index. This is a standard formulation of hierarchical policy learning, under which prior work has explored different goal representations, with $\Omega$ being the original state space~\cite{schaul2015universal} or a learned latent space~\cite{vezhnevets2017feudal}. A latent goal for $\pi^{(k)}_{L}$ can be computed using an encoder $E_k$ on the goal state, which is defined in the original state space. To harness these skills, we further design a high-level meta controller $\pi_{H}: \mathcal{S}\rightarrow \{1, 2, \ldots, K\} \times \Omega$. Intuitively, the meta controller outputs two pieces of information from a given state to invoke a low-level skill: a \revised{categorical distribution over the skill indices, of which we take the mode as the selected skill}, and a vector that specifies the goal for this selected skill to reach. With this temporal abstraction, the policy $\pi$ can be thus represented as
\begin{equation}
\label{eq:skill-composition}
\begin{split}
    \pi(a_{t}|s_{t}) = & \sum_{i=1}^{K}\mathds{1}(i=k)\pi^{(i)}_{L}(a_{t}|s_{t}, \omega), \\ & \text{ where } (k, \omega) =  \pi_{H}(s_{t})
\end{split}
\end{equation}
Prior works proposed different algorithms to train hierarchical policies with reinforcement learning~\cite{gupta2020relay} or imitation learning~\cite{mandlekar2020iris}, but typically focusing on a single task. In contrast, we examine a multi-task learning formulation, where the skills  are learned from multi-task human demonstrations. We assume the demonstrations come from $M$ different tasks, each corresponding to a different MDP. We assume the MDPs of all $M$ tasks share the same state space, action space, and transition probabilities, but differ in reward functions and initial state distributions. Our demonstrations are collected from human operators to complete instances of each task from different initial states. We denote $\mathcal{D}^{(m)}=\{\tau^{(m)}_i\}_{i=1}^{N_m}$ as the demonstration datasets for the $m$-th task, where $\tau^{(m)}_i=\{(s_t, a_t)\}_{t=0}^{T^{(m)}_i}$ is the $i$-th demonstration sequence of length $T^{(m)}_i$ and $N_m$ is the total number of demonstrations for this task. Let \yifeng{$\mathcal{D}=\bigcup_{m=1}^M \mathcal{D}^{(m)}$} be the aggregated dataset of all tasks.

Our method learns the hierarchical policies from the multi-task demonstrations. We use hierarchical clustering (Sec.~\ref{sec:clustering}) to identify the recurring temporal segments from the aggregated dataset, separating $\mathcal{D}$ into $K$ partitions $\{\tilde{\mathcal{D}}_1, \ldots, \tilde{\mathcal{D}}_K\}$. By learning on $\mathcal{D}$, we augment the training data for individual tasks and facilitate the learned low-level skills to be reusable across tasks. We use hierarchical behavior cloning (Sec.~\ref{sec:skill}) to learn the goal-conditioned skills $\pi^{(k)}_L$ on $\tilde{\mathcal{D}}_k$ for $k=1,2, \ldots, K$. Once obtaining skills, we learn to compose the skills with a task-specific meta-controller trained on the demonstration data of that task, for example, training $\pi_H$ to solve the $m$-th task on $\mathcal{D}^{(m)}$.

\begin{figure*}[t]
    \centering
    \includegraphics[width=.80\linewidth, trim=0cm 0cm 0cm 0cm,clip]{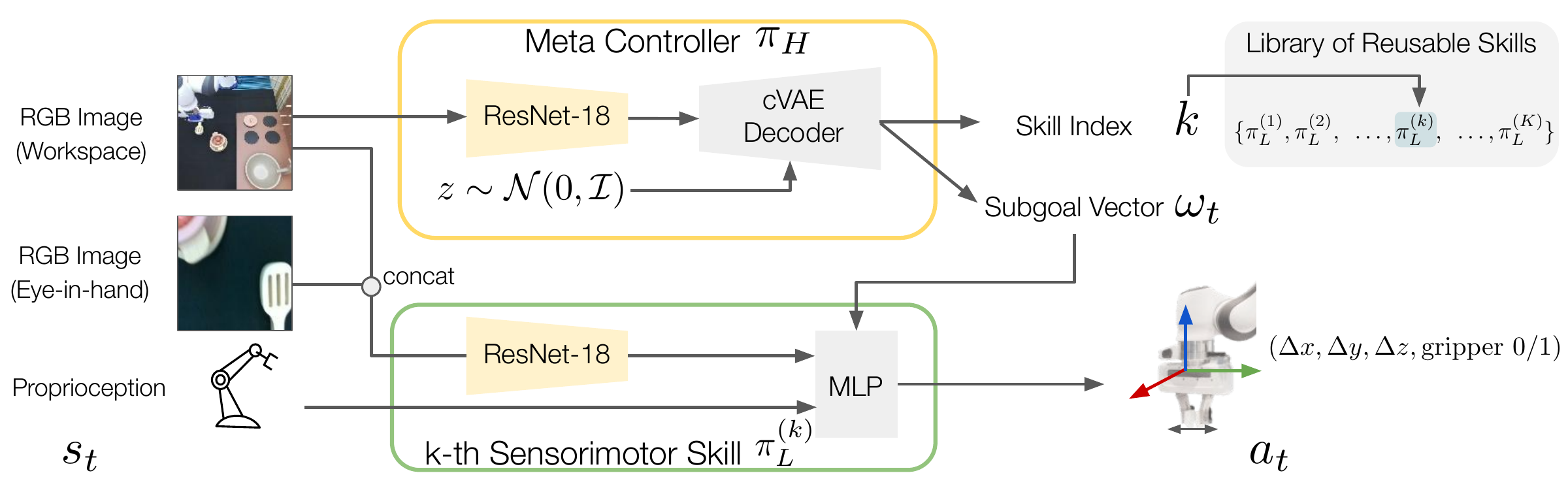}
    \vspace{-1mm}
    \caption{Overview of the hierarchical policy. Given a workspace observation, the meta controller selects the skill index and generates the latent subgoal vector $\omega_t$. Then the selected sensorimotor skill generates action $a_t$ conditioned on observed images, proprioception, and $\omega_t$. }
    \label{fig:hierarchical-policy}
    \vspace{-4mm}
%\end{minipage}
\end{figure*}

\subsection{Skill Segmentation with Hierarchical Clustering}
\label{sec:clustering}
\vspace{-2mm}
We present how to split the aggregated dataset $D$ into $K$ partitions, which we use to train the $K$ skills. Our objective is to cluster similar temporal segments from multi-task demonstrations into the same skill, easing the burden for the downstream imitation learning algorithm. To this end, we first learn per-state representations based on multi-sensory cues, which we use to form a hierarchical task structure of each demonstration sequence $\tau^{(m)}_{i}$ with bottom-up agglomerative clustering. We then identify the recurring temporal segments across the entire demo dataset via spectral clustering. The whole process is fully unsupervised without additional annotations beyond the demonstrations.

\noindent
\textbf{Learning Multi-Sensory State Representations.}
Our approach learns a latent representation per state in the demonstrations. It is inspired by research in event perception~\cite{zacks2001event}, which addresses the importance of correlation statistics presented in multiple sensory modalities for event segmentation. \ourmethod{} learns the representations from multi-modal sensory data to capture their statistical patterns. Our method follows Lee et al.~\cite{lee2020making} which learns a joint latent feature of all modalities by fusing feature embeddings from individual modalities (multi-view images and proprioception) with the Product of Experts~\cite{hinton2002training}. \yifeng{The feature is optimized over an adapted evidence lower bound loss~\cite{kingma2013auto} which reconstructs the current sensor inputs. The reconstruction is different from the one from Lee et al. which is optimized over reconstructing next states. This different design choice is motivated by the fact that the previous work focuses on policy learning with fused representation inputs which needs to encode future state information, while ~\ourmethod{} focuses on learning the statistical patterns of multi-sensory data at the current state.} By learning the joint representation, it captures the congruence underlying multi-sensory observations while retaining the information necessary to decode $s_t$. We denote $h_t$ as the latent vector computed from $s_t$.

%\revised{\yifeng{Concretely, the fused representation is optimized over an adapted evidence lower bound loss (See the implementation details from our code repository)}. A major difference is that we optimize over reconstructing the current states while Lee et al. optimize over reconstructing the next states. This different design choice is motivated by the fact that their work focuses on using the latent representation as inputs for policy learning which needs to encode future state information, while ~\ourmethod{} focus on learning the statistical patterns of multi-sensory data at the current state.}

\noindent
\textbf{Discovering Temporal Segments.} 
\ourmethod{} uses the per-state representations to effectively group states in temporal proximity to build a hierarchical representation of a demonstration sequence. The strength of a hierarchical representation, as opposed to flat segmentation, is the flexibility to decide the segmentation granularity for imitation learning. Here we use hierarchical aggomerative clustering, where in each step we combine two adjacent temporal segments into one based on similarity until all segments are combined into the entire demonstration sequence. This process produces a tree of segments. To reduce the tree depth, we start with the bottom-level elements that contain a temporal segments of a demonstration of $W$ steps ($W=10$ in our case).
The clustering process selects two adjacent segments that are most similar to each other among all pairs of adjacent segments, and merges them into one longer segment. 
The similarity between two segments is computed according to the $\ell_2$ distance between their \textit{segment features}, defined to be the average of latent vectors $\{h_t\}$ of all states in each segment. The process is repeated until we  have only one segment left for each $\tau^{(m)}_{i}$. We discover a collection of intermediate segments, which we term as \textit{temporal segments}, from the formed hierarchies. This concept is inspired by the concept of Mid-level Action Elements~\cite{lan2015action} in the action recognition literature. The way we determine the temporal segments is to breadth-first search from the root node of the hierarchy. During the search, we stop on one branch if the length of the intermediate segment is not longer than a given threshold of minimum length. And the whole breadth-first search is stopped when we have the number of segments at the lowest levels on every branch are more than a given threshold, and each segment at the lowest levels on every branch in  $\tau^{(m)}_{i}$ is a temporal segment.

\noindent
\textbf{Partitioning Skill Datasets.}
After we have a set of temporal segments for every $\tau^{(m)}_{i}$, we aggregate them from all demonstrations into one set, and apply another clustering process to group them into $K$ partitions $\{\tilde{\mathcal{D}}_{1}, \ldots, \tilde{\mathcal{D}}_{K}\}$, and we use each partition $\tilde{\mathcal{D}}_{k}$ to train the skill $\pi^{(k)}_{L}$. By training the skills on datasets from multiple tasks, it improves the reusability of the skills. We use spectral clustering~\cite{von2007tutorial} with RBF kernel on the features of temporal segments. The feature of a segment is computed as the concatenation of  representations of the first, middle (or several frames in the middle), and last states of the segment. The number of keyframes chosen in the middle of a segment can vary based on the average length of demonstrations. The spectral clustering step results in $K$ datasets of temporal segments for skill learning. 
In practice, we set the maximum number of clusters in spectral clustering and merge any classified skill into an adjacent skill if its average length is below a threshold. The number of remaining classes is denoted as $K$, which is the final number of skills we partition demonstrations into. 

\vspace{-2mm}
\subsection{Policy Learning with Hierarchical Behavioral Cloning}
\label{sec:skill}
\vspace{-2mm}
We use the obtained $K$ datasets of temporal segments from the segmentation step to train our hierarchical policies % to solve manipulation tasks, 
using a hierarchical behavioral cloning algorithm, including two parts: 1) skill learning with goal-conditioned imitation; 2) skill composition with a meta controller. Figure~\ref{fig:hierarchical-policy} visualizes the model structure of the hierarchical policy. 

\noindent
\textbf{Skill Learning with Goal-Conditioned Imitation.} 
We train each skill $\pi^{(k)}_{L}$ on the corresponding dataset $\tilde{\mathcal{D}}_{k}~\forall k=1, \ldots, K$. Every skill $\pi^{(k)}_{L}(a_t|s_t, \omega_t)$ takes a sensor observation $s_t \in \mathcal{S}$ and a subgoal vector $\omega_t$ as input, and produces a robot's motor action $a_t\in\mathcal{A}$. By conditioning $\pi^{(k)}_{L}$ on a subgoal vector, we enable the meta controller to invoke the skills and specify the subgoals that these skills should achieve.
Instead of defining the subgoals in the original sensor space, which is typically high-dimensional, we instead learn a latent space of subgoals $\Omega$, where a subgoal state $s_g$ is mapped to a low-dimensional feature vector $\omega_t \in \Omega$. For each state $s_t$, we define its subgoal as the future state either $H$ steps ahead of $s_t$ in the demonstration or the last state of a skill segment if it reaches the end of the segment from $s_t$ within $H$ steps. The reason we define a subgoal as a look-ahead state a constant number of steps in the future, as opposed to the final goal state of the task, is to exploit the temporal abstraction and reduce the computational burden of individual skills --- skills only need to reach short-horizon subgoals, without the need for reasoning about long-term goals.
Concretely, we train such a goal-conditioned skill on skill segments in $\tilde{\mathcal{D}}^{(k)}$. For each  $\pi^{(k)}_{L}$, we sample $(s_{t}, a_{t}, s_{g_{t}}) \sim \tilde{\mathcal{D}}^{(k)}$ where $g_{t}=\min{(t+H, T)}$ ($T$ is the last timestep of end of the segment),
and we generate a latent subgoal vector $\omega_{t}$ for the subgoal of $s_t$, $s_{g_{t}}$, using a subgoal encoder $\omega_{t}=E_{k}(s_{g_{t}})$, where $E_{k}(\cdot)$ is a \yifeng{ResNet-18 backboned} network jointly trained with the policy $\pi^{(k)}_{L}$. 

\noindent
\textbf{Skill Composition with A Meta Controller.}
\label{sec:meta-controller}
Now we that have a set of skills, we need a meta controller to decide which skill to use at $s_t$ and specify the desired subgoal for it to reach. We train a task-specific meta controller $\pi_{H}$ for each task. Given the current state, $\pi_{H}$ outputs an index $k\in\{1, \ldots, K\}$ to select the $k$-th skill, along with a subgoal vector $\omega_t$ on which the selected skill is conditioned. 
As human demonstrations are diverse and suboptimal by nature, the same state could lead to various subgoals in demonstration sequences (e.g., grasp different points of an object, push an object at different contact points). Thus, the meta controller needs to learn distributions of subgoals from demonstration data of a task $\mathcal{D}^{(m)}$, and we choose conditional Variational Autoencoder (cVAE)~\cite{kingma2013auto}. To obtain the training data of skill indices and subgoal vectors, we sample $(s_t, s_{g_t})\sim \mathcal{D}^{(m)}$, and from the clustering step we have the correspondence between a skill index $k$ and state $s_t$ while from the skill learning step we can generate a per-state subgoal vector $\omega_{t}=E_{k}(s_{g_{t}})$. The meta controller $\pi_{H}$ is trained to generate a skill index $k$ and a subgoal vector $\omega_{t}$ conditioned on state $s_t$. During evaluation, the controller generates the skill index and the subgoal vector conditioned on the current state $s_t$ and a latent vector $z$ from the prior distribution $\mathcal{N}(0, \mathcal{I})$. The controller for evaluation is typically chosen to operate at a lower frequency than skills so that it can avoid switching among skills too frequently.

\revised{Training meta controllers follows the same cVAE training convention in prior works~\cite{mandlekar2020iris, mandlekar2020learning} which minimizes an ELBO loss on demonstration data. To obtain the training supervision for skill indices and subgoal vectors, we augment the demonstrations with results from the clustering and skill learning steps: 1) The training labels of skill indices come from the cluster assigments; 2) The latent subgoal vectors are computed on the demonstration states, and the encoders for computing the vectors were jointly trained with skill policies.}
\vspace{-6mm}

\section{Experiments}
\label{sec:experiments}
\vspace{-3mm}
We design our experiments to examine three questions: 1) How does \ourmethod{} perform in long-horizon vision-based manipulation tasks compared to baseline methods? 2) Do learning from multi-task demonstrations and using multimodal features improve the quality and reusability of skills? and 3) Does~\ourmethod{} work with real-world sensor data and physical hardware?

\begin{figure}[h]
\centering
\begin{minipage}[t]{0.17\linewidth}
        \includegraphics[width=\linewidth,trim=0cm 0cm 0cm 0cm,clip]{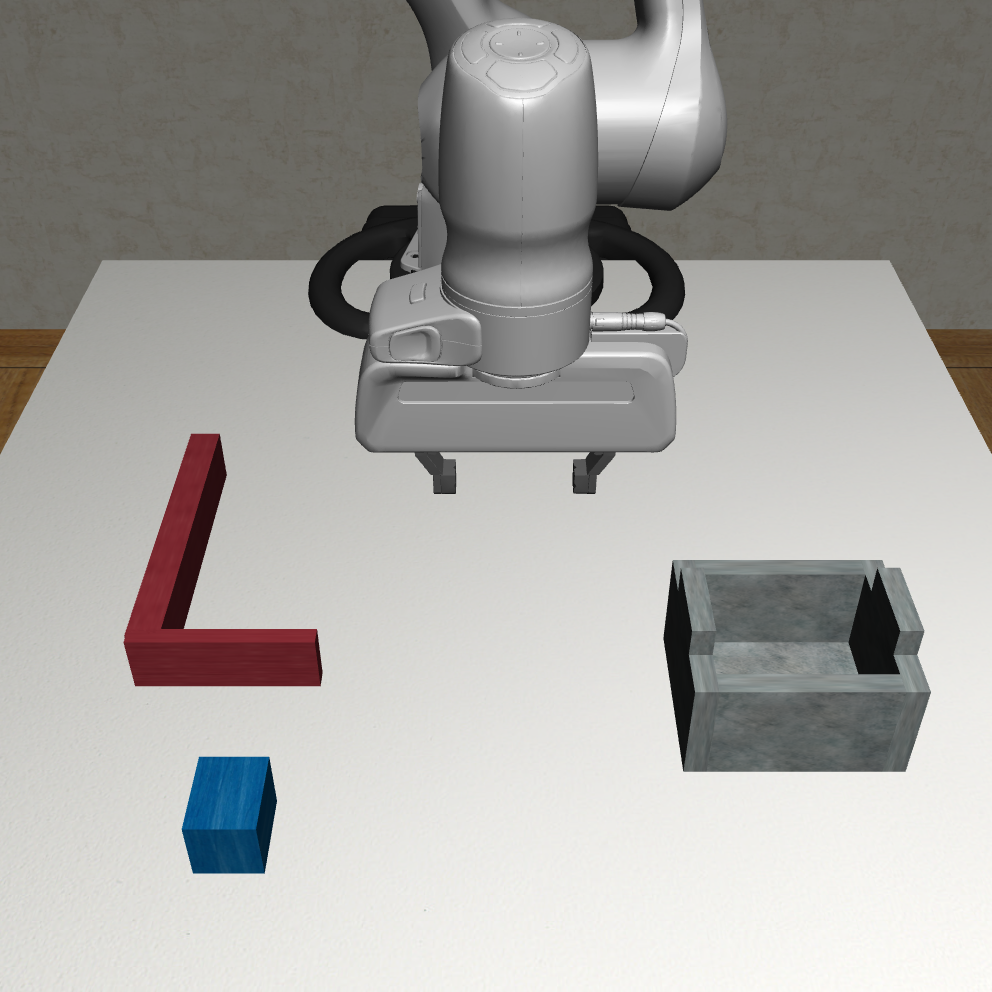}
        \vspace{-5mm}
     \subcaption{}
      \end{minipage}
      \hfill
 \begin{minipage}[t]{0.17\linewidth}
        \includegraphics[width=\linewidth,trim=0cm 0cm 0cm 0cm,clip]{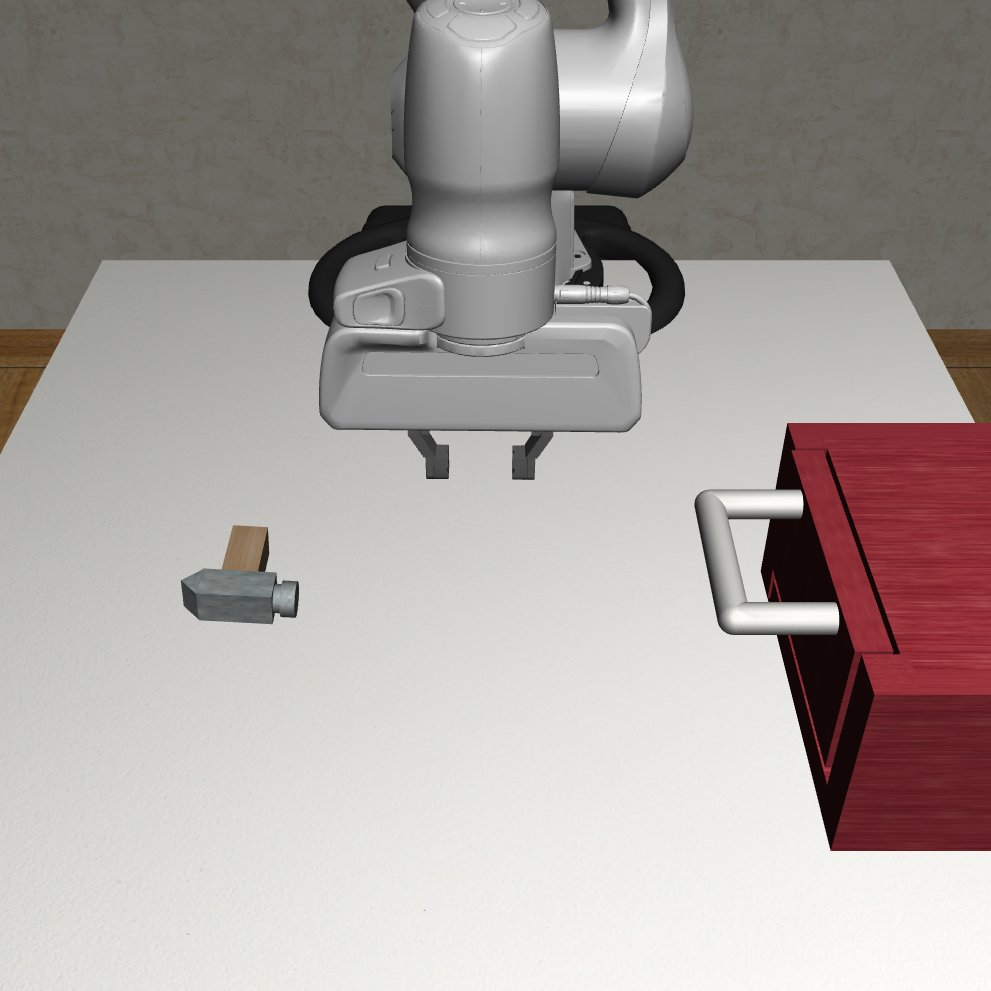}
        \vspace{-5mm}        
        \subcaption{}
      \end{minipage}
      \hfill
     \begin{minipage}[t]{0.17\linewidth}
        \includegraphics[width=\linewidth,trim=0cm 0cm 0cm 0cm,clip]{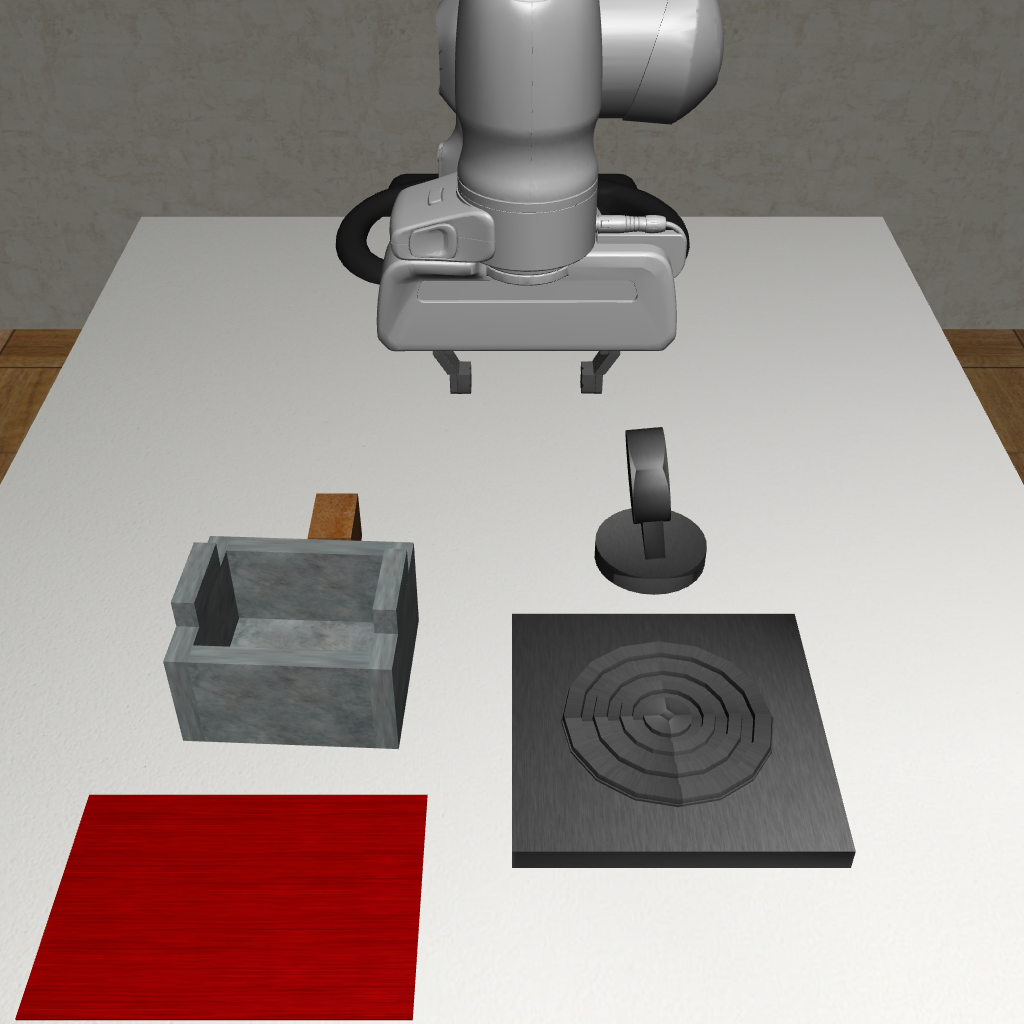}
        \vspace{-5mm}        
    \subcaption{}
      \end{minipage}   
      \hfill
     \begin{minipage}[t]{0.17\linewidth}
        \includegraphics[width=\linewidth,trim=0cm 0cm 0cm 0cm,clip]{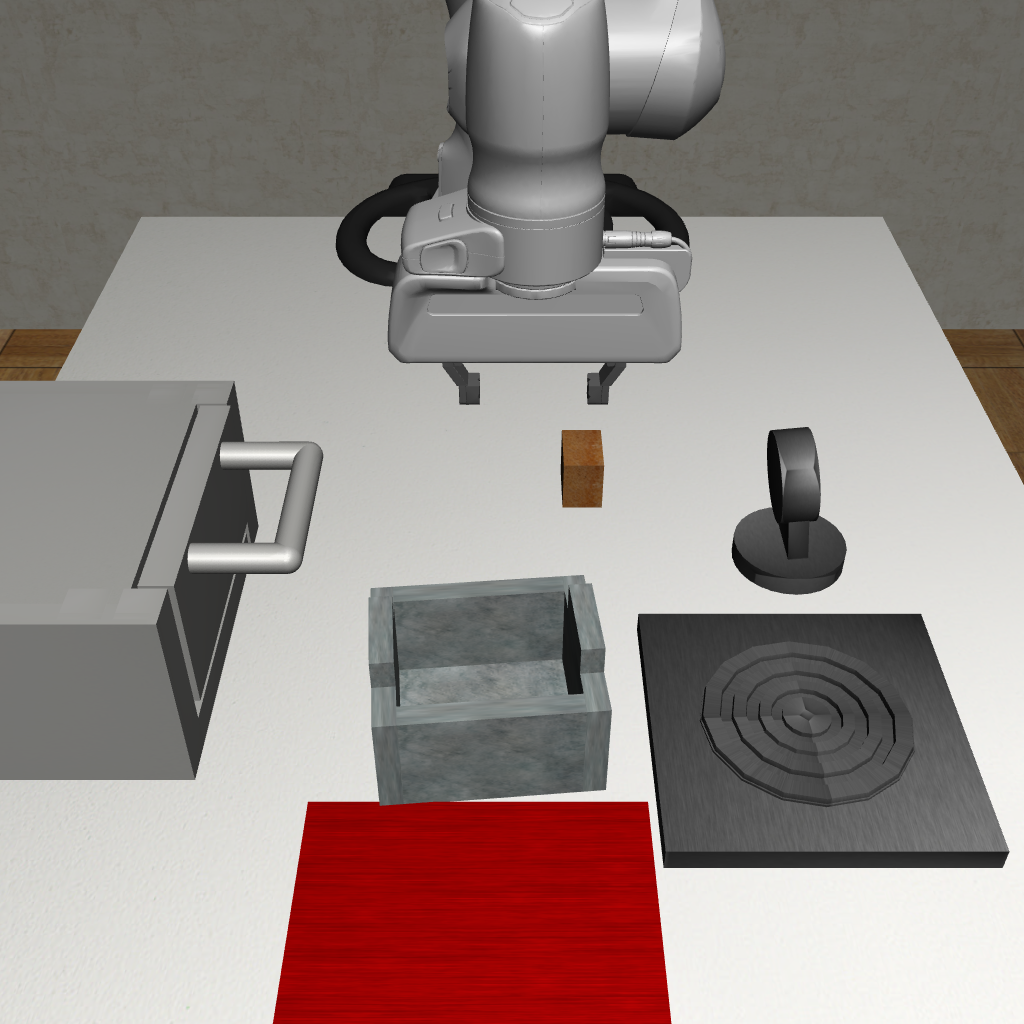}
        \vspace{-5mm}        
     \subcaption{}   
      \end{minipage}        
      \hfill
     \begin{minipage}[t]{0.17\linewidth}
        \includegraphics[width=\linewidth,trim=0cm 0cm 0cm 0cm,clip]{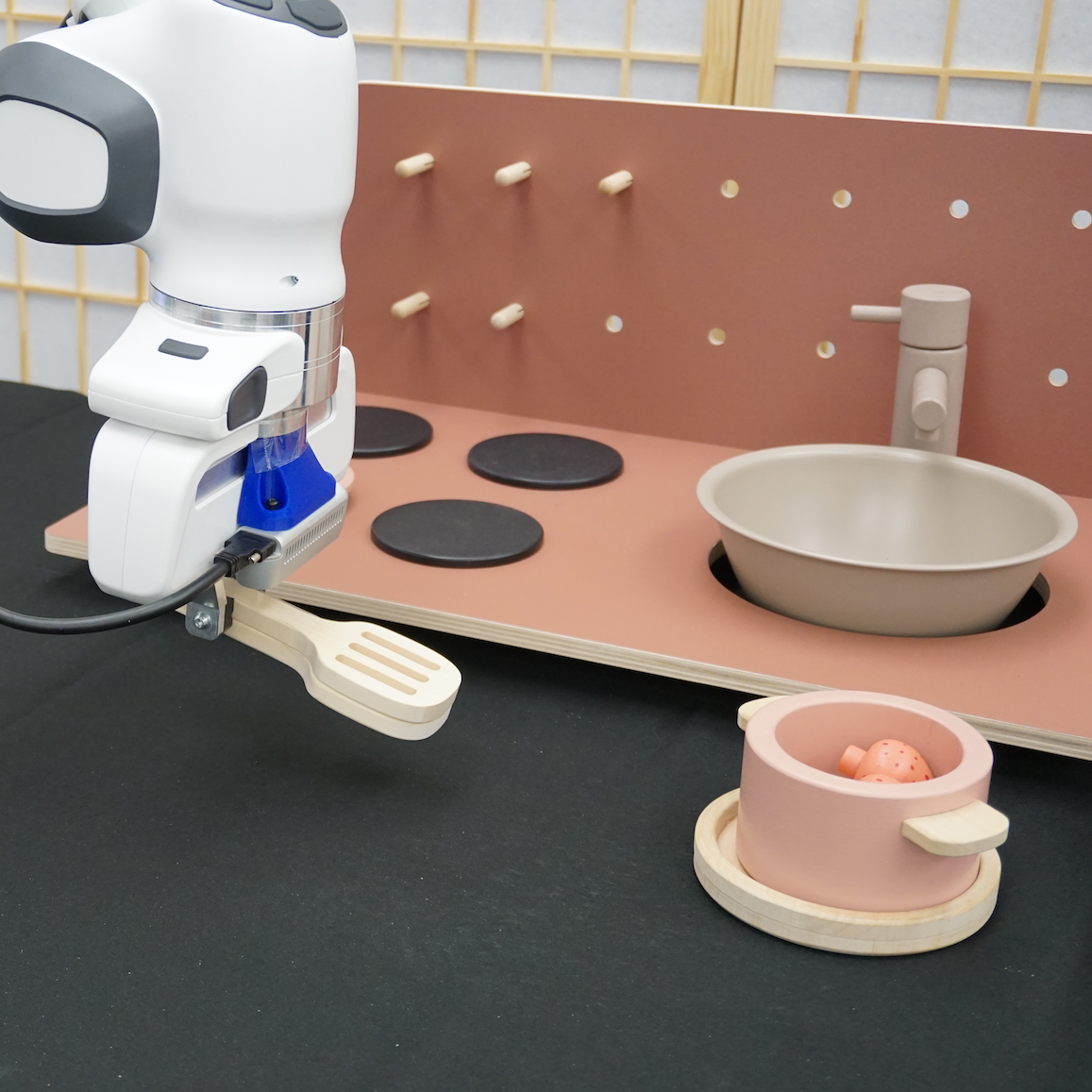}
        \vspace{-5mm}         
     \subcaption{}   
      \label{fig:real}
      \end{minipage}         
    \vspace{-1mm}
     \caption{Visual illustrations of the four simulation tasks and one real robot task used in our experiments.
    (a)~\tooluse{}; (b)~\hammer{}; (c)~\kitchen{}; (d)~\multitask{}; (e)~\realrobot{}.
    }
    \label{fig:experiment}
    \vspace{-7mm}
\end{figure}

\subsection{Experimental Setup}
\label{sec:experiment-setup}
\vspace{-3mm}
% We introduce the overview of environment designs and data collection process for our experiment results. %For more details of environments (task descriptions), please see Sec.~\ref{sec:app-env}.
We perform baseline comparisons and model analysis in simulation environments developed with the \texttt{robosuite} framework~\cite{zhu2020robosuite}, and present quantitative results on real hardware. Figure~\ref{fig:experiment} illustrates all the tasks. The first three single-task environments, \tooluse, \hammer, and \kitchen,~are designed primarily for baseline comparisons and ablation studies. The multi-task domain \multitask{}~is designed for investigating the quality and reusability of skills discovered from multi-task demonstrations. The \realrobot{} task is for real-world validation and deployment. We provide detailed task descriptions below. For all the experiments, we use a 7-DoF Franka Emika Panda arm with a position-based Operational Space Controller~\cite{khatib1987unified} and a binary command for controlling parallel-jaw gripper. The meta controller runs at $4$Hz and the skills run at $20$Hz. We release our datasets, simulation environments, and model implementations on our project website for reproducing purpose.

\noindent
\textbf{Single tasks.} The three single tasks require prolonged interactions with the environment and entail a broad range of prehensile and nonprehensile behaviors, such as tool use and manipulating articulated objects. \revised{The goal of \tooluse{} is to put the cube into the metal pot. In order to fetch the cube which is initially beyond the robot's reach, the robot needs to first grasp an L-shape tool and pull the cube with the tool. After fetching the cube, it needs to put the tool aside, pick up the cube, and place it into the pot. The goal of \hammer{} is to put the hammer in the drawer and close the drawer, where the hammer is small and hard to grasp. To achieve this goal, the robot needs to open the drawer, place the hammer into the drawer, and close the drawer. The goal of \kitchen{} is to cook and serve a simple dish in the serving region and turn off the stove. This task is the most complex among the three, requiring a sequence of subtasks, including turning on the stove, placing the pot on the stove, putting the ingredient into the pot, putting the pot on the table, pushing it to the serving region (red region on the table), and turning off the stove at the end.}

\noindent
\textbf{\multitask{}.} The multi-task domain includes three tasks (referred to as \task{1}, \task{2}, and \task{3}) that are distinct from each other by their end goals. The goal state of \task{1} entails \revised{the drawer closed, the stove on, the cube in the pot, and the pot placed on the stove. The goal of \task{2} entails the drawer closed, the stove on, the cube in the pot, and the pot in the serving region. The goal of \task{3} entails the drawer closed, the stove off, the cube in the pot, and the pot in the serving region.} To study the reusability of our skills, we design three variants for each task based on their initial configurations, which we refer to as \variant{1}, \variant{2}, \variant{3}. \revised{We describe all task variants in details  at our project website.} Different initial configurations require solving different combinations of subtasks. Therefore, we examine whether skills learned in a subset of task variants can be reused in new variants.

\noindent
\textbf{\realrobot{}.}  The task requires versatile behaviors including grasping, pushing, and tool use. The robot needs to remove the lid of the pot, place the pot to the plate, pick up the tool, use the tool to push the pot along with the plate to the side of the table, and put down the tool in the end. We capture \yifeng{RGB} images from the workspace camera (Kinect Azure) and the eye-in-hand camera (Intel Realsense D435i).

\noindent
\textbf{Data Collection.} We collect human demonstrations through teleoperation with a 3Dconnexion SpaceMouse. We collect $100$ demonstrations for each of the three single tasks (less than 30 minutes each task), $120$ demonstrations for each task ($40$ for each of the three task variants) in \multitask{}, and $50$ demonstrations for \realrobot{}. Each demonstration consists of a sequence of sensory observations (images from the workspace and the eye-in-hand camera, proprioception) and actions (the end-effector displacement and the gripper open/close).

\vspace{-2mm}
\subsection{Quantitative Results}
\vspace{-3mm}
For all simulation experiments, we evaluate \ourmethod{} and baseline methods in each task for 100 trials with random initial configurations (repeated with 5 random seeds). We use the success rate over trials as the evaluation metric, and a trial is considered successful if the task goal is reached within the maximum number of steps.

\noindent
\textbf{Single Task Experiments.} Here we compare \ourmethod{} with imitation learning baselines on the single-task environments. To examine the efficacy of hierarchical modeling for long-horizon tasks, we first compare with a Behavior Cloning (BC) baseline~\cite{zhang2018deep} which trains a flat policy on the demonstrations. To examine our bottom-up clustering-based segmentation method, we compare with a second baseline that uses a classical Change Point Detection (CP) algorithm~\cite{niekum2015online} to temporally segment the demonstrations while keeping the rest of the model design identical to ours.

Table~\ref{tab:single-task-results} reports the quantitative results. BUDS outperformed both baselines for all three tasks, by over 20\% on average. The comparison between BC and \ourmethod{} shows that while BC is able to solve short-horizon task reasonably well, it suffers a significant performance drop in longer tasks, such as \kitchen. In contrast, \ourmethod{} breaks down a long-horizon task with skill abstraction, leading to a consistent high performance across tasks of varying lengths.  The comparison between CP and \ourmethod{} suggests that the quality of skill segmentation plays an integral role in final performance. Qualitatively, we found that the CP baseline failed to produce coherent segmentation results across different demonstrations, hindering the efficacy of policy learning.

We observe two major failure modes in \ourmethod{}: 1) Incorrect selection of skills due to out-of-distribution states, 2) Manipulation failures due to imprecise grasps. \revised{We quantify the failure modes in the \kitchen{} task with 5 repeated runs. Failures due to the first mode take up $12.3\% \pm 2.9\%$ of the evaluation trials, and failures due to the second one take up $9.0\% \pm 3.6\%$.} Both failure types pertain to the fundamental limitations of imitation learning on small offline datasets. We believe the model performance could be improved with large-scale training and online robot experiences. We leave it for future work.

\begin{table}[t]
\centering
\vspace{3mm}
\caption{\label{tab:single-task-results} Success rate (\%) in single task environments.}
%\vspace{-2mm}
\makeatletter\def\@captype{table}
  \resizebox{0.8\linewidth}{!}{  
  \begin{tabular}{lccc}
    \toprule
    \textbf{Environments} & \textbf{BC~\cite{zhang2018deep}} & \textbf{CP~\cite{niekum2015online}} & \textbf{\ourmethod (Ours)}\\
    \midrule
    \tooluse~ & 54.0 $\pm$ 6.3 & 36.8 $\pm$ 5.1 & \textbf{58.6} $\pm$ 3.1  \\
    \hammer~   & 47.8 $\pm$ 3.7 & 60.4 $\pm$ 4.5 & \textbf{68.6} $\pm$ 5.7 \\
    \kitchen~  & 24.4 $\pm$ 5.3 &  23.4 $\pm$ 3.4   & \textbf{72.0} $\pm$ 4.0\\
    \bottomrule
  \end{tabular}
  }
\vspace{-0.75cm}
% \caption{Sample table title}
\end{table}

\noindent
\textbf{Comparisons to Hierarchical Imitation Learning Algorithms%~\cite{mandlekar2020learning,gupta2020relay,mandlekar2020iris}
.} 
BUDS shares the same principle with recent works on hierarchical imitation learning, including IRIS~\cite{mandlekar2020iris}, GTI~\cite{mandlekar2020learning}, and RPL~\cite{gupta2020relay}. One notable distinction is that the prior works consider a single low-level skill, rather than a library of skills. 

To compare BUDS with GTI, we evaluate our method with varying numbers of skills through a parameter sweep on the number of clusters $K$ in the spectral clustering step. In the special case when \ourmethod{} has only a single skill ($K=1$), our method is equivalent to a variant of GTI without the image reconstruction term.
Table~\ref{tab:varying-K} reports the results in the \kitchen{} task. We observe that the number of skills has a salient impact on model performance. Intuitively, when $K$ is too small, each skill will have difficulty dealing with diverse subgoals and various visual observations; and when $K$ is too large, each skill has fewer data points to train on, as the dataset is fragmented into smaller partitions. The peak performance is observed with $K=6$ skills, which is the value we used for the main experiments.

The GTI variant with a single skill ($K=1$) fails to achieve non-zero task success. We also implemented the original GTI with the image reconstruction term, but observed no significant change in performance. After analyzing the qualitative behaviors of the GTI policy, we find that it works fine if the initial state is close to the task goal, but it cannot handle initial states that are further away. For quantitative evidence, we conduct an additional evaluation with the \tooluse{} task, where we reset the robot to the state when it has already fetched the cube and placed the tool down. To complete the task, the robot only needs to pick up the cube and place it in the pot. In this shorter subtask, the GTI variant and BUDS achieved $63.0\%$ and $60.3\%$ success rates respectively. In comparison, they achieved $0.0\%$ and $58.6\%$ (Table~\ref{tab:single-task-results}) success rates when starting from the original initial states. These results imply that GTI does not generalize well to longer tasks studied in this work. 

Comparisons to IRIS and RPL require additional efforts as they were designed for low-dimensional states. We adapt RPL to handle image inputs by extracting visual features with a ResNet-18 module in the policy network, but the adapted model achieves no task success. After a closer examination, RPL fails to generalize to various object placements. Furthermore, it also uses a single low-level policy similar to GTI, which we have shown lower performance with the GTI-variant. On the other hand, IRIS is more difficult to adapt as its high-level policy predicts subgoals in the original state space, in our case, the raw sensory space. We expect it to suffer the similar issues as the other two methods.

\begin{table}[t!]
\centering
\vspace{3mm}
    \caption{\label{tab:varying-K} Results in \kitchen{} with varying numbers of skills}  
%\vspace{-1mm}
\makeatletter\def\@captype{table}
  \resizebox{1.0\linewidth}{!}{  
  \begin{tabular}{lccccc}
    \toprule
    \textbf{} & $K=1$ (GTI~\cite{mandlekar2020learning}) & $K=3$ & $K=6$ & $K=9$ & $K=11$ \\
    \midrule
    \kitchen~ & $0.0\pm0.0$ & $24.2\pm3.6$ & $72.0\pm4.0$ & $60.6\pm6.53$ & $44.6\pm3.38$  \\
    \bottomrule
  \end{tabular}
  }
  \vspace{-2mm}
  \end{table}

\noindent
\textbf{Learning from Multi-task Demonstrations.}
\label{sec:multitask}
We investigate if \ourmethod{} is effective in learning from multi-task demonstrations. ~\ourmethod{} discovers $K=8$ skills in the \multitask{} domain, and we examine the skills from two aspects: 1) \textit{quality}: are skills learned from multi-task demonstrations better than those from individual tasks? 2) \textit{reusability}: can these skills be composed to solve new task variants that require different subtask combinations? 

We evaluate three settings: 1) \trainmulti: the skills are discovered and trained on the multi-task demonstrations, and the meta controller is trained for each task respectively; 2) \trainsingle: the skills are discovered from demonstrations of each individual task\yifeng{;} so is the meta-controller; and 3) \test: the skills are trained on demonstrations of \variant{1} and \variant{2} and the meta-controller is trained on \variant{3}. Table~\ref{tab:multitask} presents the evaluation results. The comparisons between \trainmulti{} and \trainsingle{} indicate that skills learned across multi-task demonstrations improve the average task performance by $8\%$ compared to those learned on demonstrations of individual tasks. We hypothesize that the performance gain roots from our method's ability to augment the training data of each skill with recurring patterns from other tasks' demonstrations. The results on \test{} show that we can effectively reuse the skills to solve the new task variants that require different combinations of the skills by solely training a new meta controller to invoke the pre-defined skills. \yifeng{We also observe the low performance of \variant{3} on \test{}, because it has more subtasks than its training counterparts, and the execution failure of each skill compounds, leading to the low success rate.} We provide additional visualizations of our skills and policy rollouts on our project website.

\begin{table}[t]
\centering
\caption{\label{tab:multitask} Success rate (\%) in \multitask{}.}
\vspace{0mm}
\makeatletter\def\@captype{table}
 \resizebox{0.75\linewidth}{!}{  
  \begin{tabular}{lccc}% {lllll}
    \toprule
    \textbf{} & \trainmulti{} & \trainsingle{} & \test{}\\
    \midrule
    \task{1} &  $70.2 \pm 2.2$ & $52.6 \pm 5.6$ & $59.0 \pm 6.4$ \\
    \task{2} & $59.8 \pm 6.4$  & $60.8 \pm 1.9$ &  $55.3 \pm 3.3$  \\
    \task{3} &$75.0 \pm 2.0$ & $67.6 \pm 1.8$  & $28.4  \pm 1.5$ \\    
    % \trainsingle{} & $52.6 \pm 5.6$ & $60.8 \pm 1.9$ & $67.6 \pm 1.8$ \\
    % \trainmulti{} & $70.2 \pm 2.2$ & $59.8 \pm 6.4$ & $75.0 \pm 2.0$  \\
    % \test{} & $59.0 \pm 6.4$ & $55.3 \pm 3.3$ & $28.4  \pm$ 1.5 \\
    \bottomrule
  \end{tabular}
  }
  \vspace{-0.3cm}
\end{table}

\begin{table}[h]
\centering
\caption{\label{tab:ablation-study} Ablation study on demonstration state representations.}
\vspace{-2mm}
\makeatletter\def\@captype{table}
   \resizebox{1.0\linewidth}{!}{  
  \begin{tabular}{lcccc}
    \toprule
    \textbf{} & \textbf{\ourmethod} & \textbf{\ourmethodimage} &
    \revised{\textbf{\ourmethodwsimage}} &
    \textbf{\ourmethodproprio}\\
    \midrule
    \kitchen~ & $72.0\pm4.0$ & $41.4\pm2.2$ & \revised{$7.4\pm2.4$} & $36.8\pm5.8$  \\
    \bottomrule
  \end{tabular}
  }
  \vspace{-0.35cm}
\end{table}

\begin{figure}[t]
% \begin{minipage}[t]{0.4\linewidth}
\centering
% \end{minipage}
    \begin{minipage}[t]{1.0\linewidth}
        \includegraphics[width=\linewidth,trim=0cm 0cm 0cm 0cm,clip]{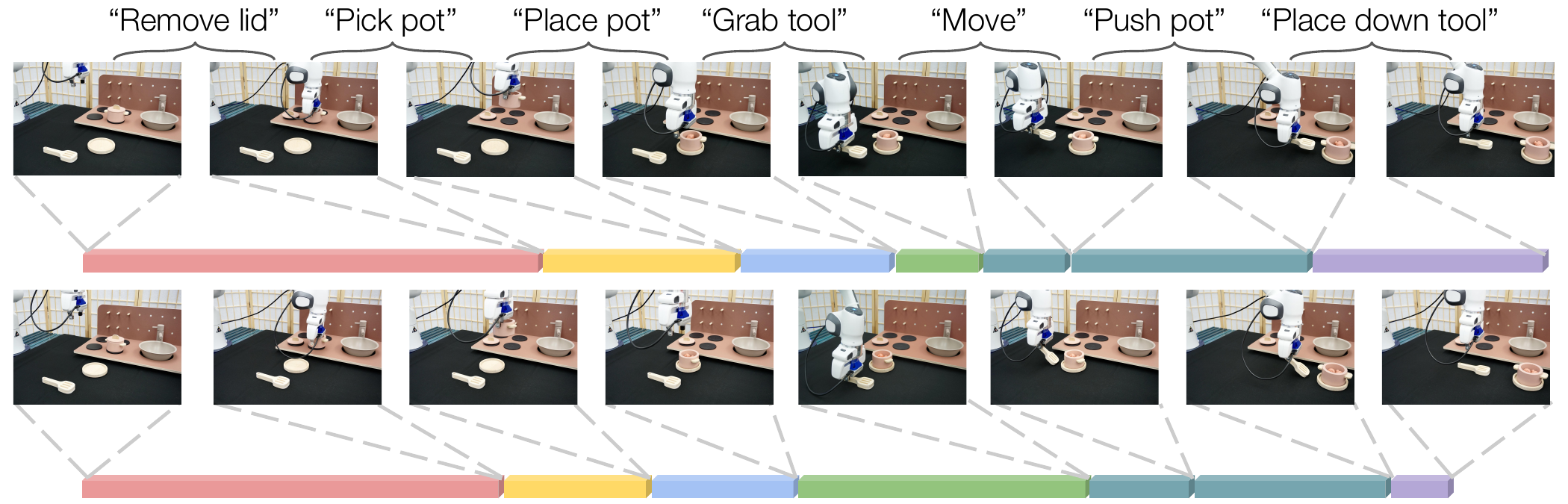}
    %   \subcaption{Hierarchical Task Structure. Red frames correspond to mid-level segments in this sequence.}
    \end{minipage}
    % \vspace{-2mm}
    \caption{Visualization of temporal segments in two demonstrations of \realrobot{} with different lengths. Each cuboid corresponds to a temporal segment, and each color represents one skill that the temporal segment is clustered into. Text annotations are our interpretations of the segmented skills. We show that the temporal segments, though discovered without supervision, nicely capture semantically meaningful subtasks. They are consistent across demonstrations despite the differences in motions, e.g., the states of the tool during pushing are very different from each other.}
    \label{fig:skill-segment}
    \vspace{-4mm}
\end{figure}

\noindent
\textbf{Ablation Study on Demonstration State Representations.} The quality of segmentation heavily relies on the choice of features we use to represent the demonstration data. A critical design of \ourmethod{} is to use multimodal representations learned from multi-view images and proprioceptive data for each state. This ablation study analyzes its impact. 
We compare \ourmethod{} with three ablative models, which learns the state representations from both workspace and eye-in-hand cameras (\ourmethodimage{}), from only the workspace camera (\ourmethodwsimage{}), and from the proprioceptive data (\ourmethodproprio{}). The rest of these ablative models remains identical to \ourmethod{}. Table~\ref{tab:ablation-study} reports the comparisons. The use of multimodal representations in \ourmethod{} substantially outperforms the ablative models in task success rate. \revised{We note that the ablative model \ourmethodwsimage{} results in significantly lower performance. The workspace images without the aid of eye-in-hand images do not capture all task-relevant information of objects due to occlusion, leading to the poor segmentation results.} This study shows that the use of multi-sensory observation leads to a more coherent task structure and better skill learning for solving challenging manipulation tasks. 

\noindent
\textbf{Real Robot Experiments.}
We perform evaluations in the \realrobot{} task to validate the practicality of \ourmethod{} for solving real-world manipulation tasks. Quantitatively, we evaluate $50$ trials on varying initial configurations, achieving a $56 \%$ success rate. The performance is at the same level as our simulation evaluations, showing that \ourmethod~generalizes well to real-world data and physical hardware. \revised{We also evaluate the most competitive baseline \textbf{CP} model on the real robot, which only achieved a $18 \%$ success rate. A consistent failure mode of this baseline is that the robot failed to place the pot correctly on the plate.}  We also qualitatively visualize the temporal segments of two demonstration sequences collected for the \realrobot{} task in Figure~\ref{fig:skill-segment}. While our clustering-based segmentation algorithm is fully unsupervised, our quantitative inspection identifies consistent segments that can be interpreted with semantic meanings.

\section{Conclusion}
\label{sec:conclusion}
\vspace{-2mm}
% In this paper, we present an unsupervised approach to discover skills from raw sensorimotor space in a bottom-up manner. Unlike existing works, our method can directly discover skills without having a model of objects a priori. 

We presented \ourmethod{}, a hierarchical approach to tackling vision-based manipulation by discovering sensorimotor skills from unsegmented demonstrations. \ourmethod{} identifies recurring patterns from multi-task  human demonstrations based on multi-sensory cues. Then it trains the skills on the recurring temporal segments with imitation learning and design a meta controller to compose these skills for tasks. The results show the effectiveness of \ourmethod{} in simulation and on real hardware. We also examine the impacts of different model designs through ablation studies.

% We show that \ourmethod{} is capable of learning closed-loop sensorimotor policies for  using a hierarchical behavioral cloning approach both in simulation and on real hardware. 

While \ourmethod{} achieved superior performances over baselines in our evaluation tasks, it suffers from the common limitations of learning from offline demonstration datasets. One future direction is to improve its performance with the robot's online experiences. Another limitation of our current approach is the need of task-specific meta controllers to compose the skills for individual tasks. For future work, we would like to develop planning methods that integrate these acquired skills with a high-level task planner, such that they can compose the skills to solve novel manipulation tasks without training a new meta controller.

% This requires the emergence of state abstraction that entailed by sensorimotor skills~\cite{konidaris2018skills,konidaris2019necessity}, but our approach can serve as the first step before discovering object-centric abstraction from actions.

% Future work will include how to combine this bottom-up framework with a top-down framework~\cite{zacks2001event}, such that we can equip the robot with a full-fledged ability to understand the events  underlying demonstrations.
%\section*{Acknowledgement}
%This work has taken place in the Robot Perception and Learning Group (RPL) and Learning Agents Research Group  (LARG)  at  UT  Austin.  RPL research has been partially supported by NSF CNS-1955523, the MLL Research Award from the Machine Learning Laboratory at UT-Austin, and the Amazon Research Awards. LARG  research  is  supported in  part  by  NSF  (CPS-1739964,  IIS-1724157,  NRI-1925082),ONR  (N00014-18-2243),  FLI  (RFP2-000),  ARO  (W911NF-19-2-0333),   DARPA,   Lockheed   Martin,   GM,   and   Bosch.Peter  Stone  serves  as the Executive Director of Sony AI America and receives financial compensation for this work. The terms of this arrangement have been reviewed and approved by the University of Texas at Austin in accordance with its policy on objectivity in research.

\section*{Acknowledgement}
\vspace{-2.5mm}
This work has taken place in the Robot Perception and Learning Group (RPL) and Learning Agents Research Group  (LARG)  at  UT  Austin. RPL research has been partially supported by NSF CNS-1955523, the MLL Research Award from the Machine Learning Laboratory at UT-Austin, and the Amazon Research Awards. LARG  research  is  supported in  part  by  NSF  (CPS-1739964,  IIS-1724157,  NRI-1925082), ONR  (N00014-18-2243),  FLI  (RFP2-000),  ARO  (W911NF-19-2-0333), DARPA,   Lockheed Martin, GM, and   Bosch. Peter Stone serves as the Executive Director of Sony AI America and receives financial compensation for this work. The terms of this arrangement have been reviewed and approved by the University of Texas at Austin in accordance with its policy on objectivity in research.

% \clearpage
\vspace{-2mm}
\balance
\printbibliography
% \clearpage
% \newpage
% \newcommand{\todocitation}{\textcolor{red}{[citation]}}
% \newcommand{\todo}[1]{\textcolor{red}{TODO: #1}}
% \newcommand{\yifeng}[1]{\textcolor{blue}{Yifeng: #1}}
% \newcommand{\yuke}[1]{\textcolor{red}{Yuke: #1}}
% \newcommand{\question}[1]{\textcolor{red}{#1}}
% \newcommand{\change}[1]{\textcolor{blue}{#1}}
% \appendix
% \input{supp}

\end{document}